\documentclass[runningheads]{llncs}
\usepackage{graphicx}
\usepackage{amsmath,amssymb} 
\usepackage{color}
\usepackage{booktabs}
\usepackage{url}

\usepackage
[compatibility=false]
{caption}
\usepackage[list=true]{subcaption}

\begin{document}

\newcommand{\RS}[1]{\textcolor{blue}{#1}}

\title{Segmenting Medical Instruments in Minimally Invasive Surgeries using AttentionMask} 
\titlerunning{Segmenting Medical Instruments in MIS Images using AttentionMask} 


\author{Christian Wilms\inst{1} \and Alexander Michael Gerlach\inst{1} \and R\"udiger Schmitz\inst{2} \and \\ Simone Frintrop\inst{1}}

\institute{University of Hamburg, Department of Informatics\\
\email{\{christian.wilms,alexander.gerlach,simone.frintrop\}@uni-hamburg.de} \and
University Medical Center Hamburg-Eppendorf\\
\email{r.schmitz@uke.de}}
%

\authorrunning{C. Wilms et al.} 



\maketitle

\begin{abstract}
Precisely locating and segmenting medical instruments in images of minimally invasive surgeries, medical instrument segmentation, is an essential first step for several tasks in medical image processing.  However, image degradations, small instruments, and the generalization between different surgery types make medical instrument segmentation challenging. To cope with these challenges, we adapt the object proposal generation system AttentionMask and propose a dedicated post-processing to select promising proposals. The results on the ROBUST-MIS Challenge 2019 show that our adapted AttentionMask system is a strong foundation for generating state-of-the-art performance. Our evaluation in an object proposal generation framework shows that our adapted AttentionMask system is robust to image degradations, generalizes well to unseen types of surgeries, and copes well with small instruments.

\end{abstract}

\section{Introduction}
\label{sec:intro}
\begin{figure}[t]
\centering
\begin{tabular}{ccc}
\includegraphics[width = .31\linewidth]{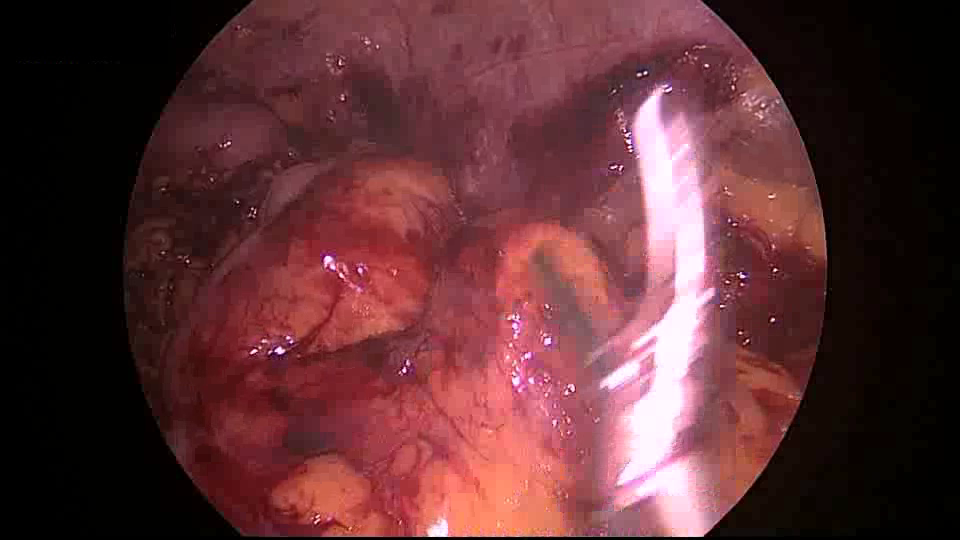} &
\includegraphics[width = .31\linewidth]{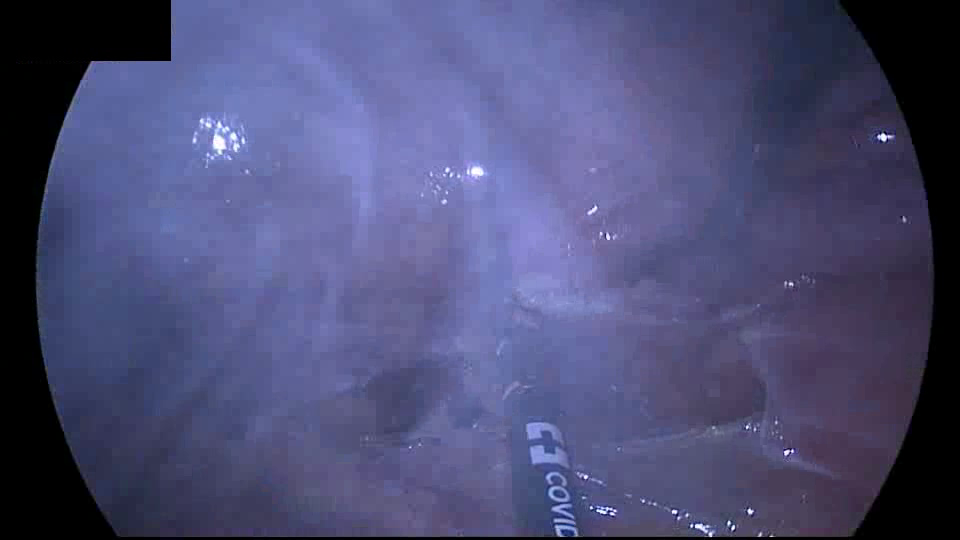} &
\includegraphics[width = .31\linewidth]{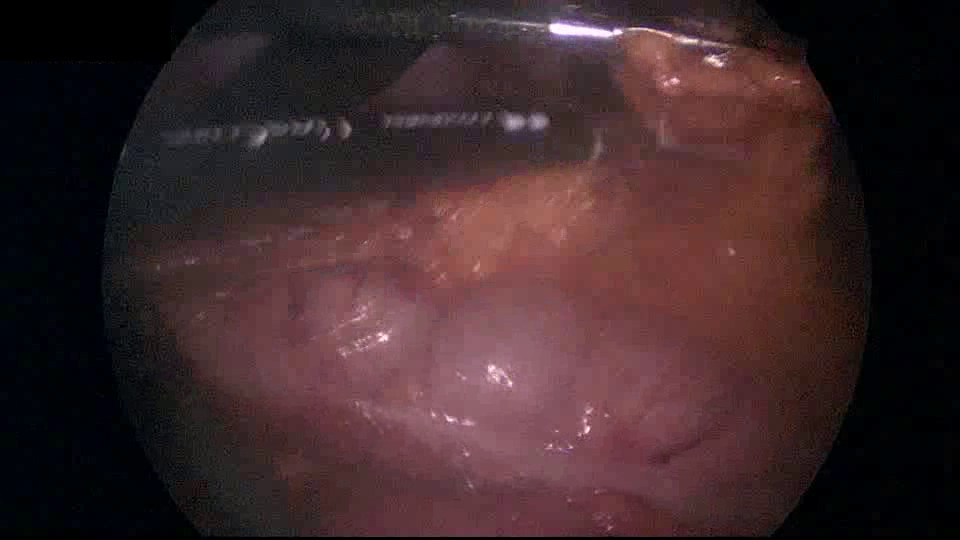}\\
\includegraphics[width = .31\linewidth]{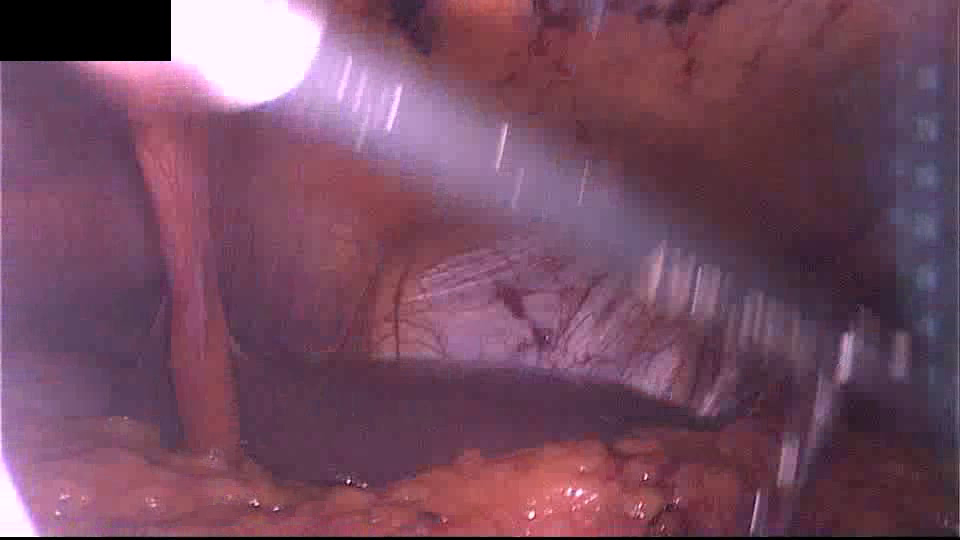} &
\includegraphics[width = .31\linewidth]{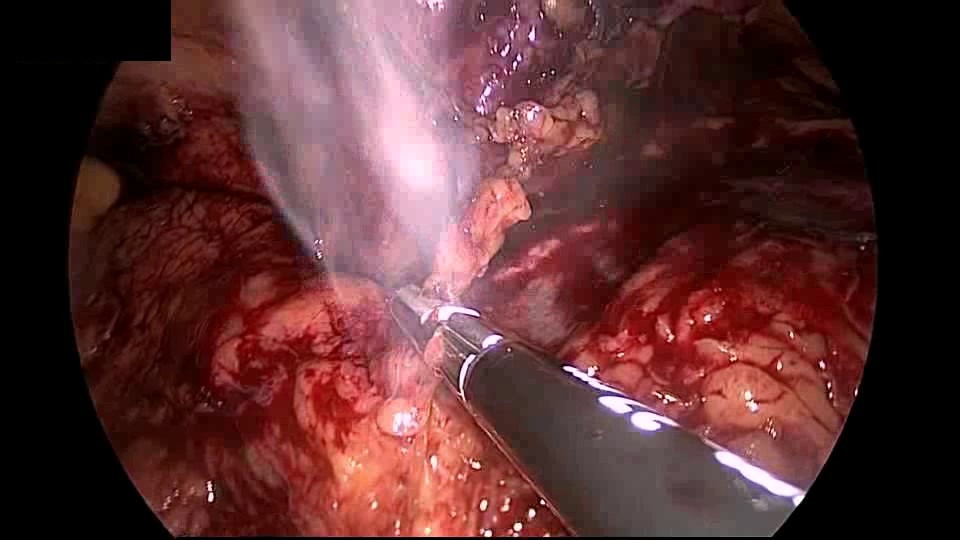} &
\includegraphics[width = .31\linewidth]{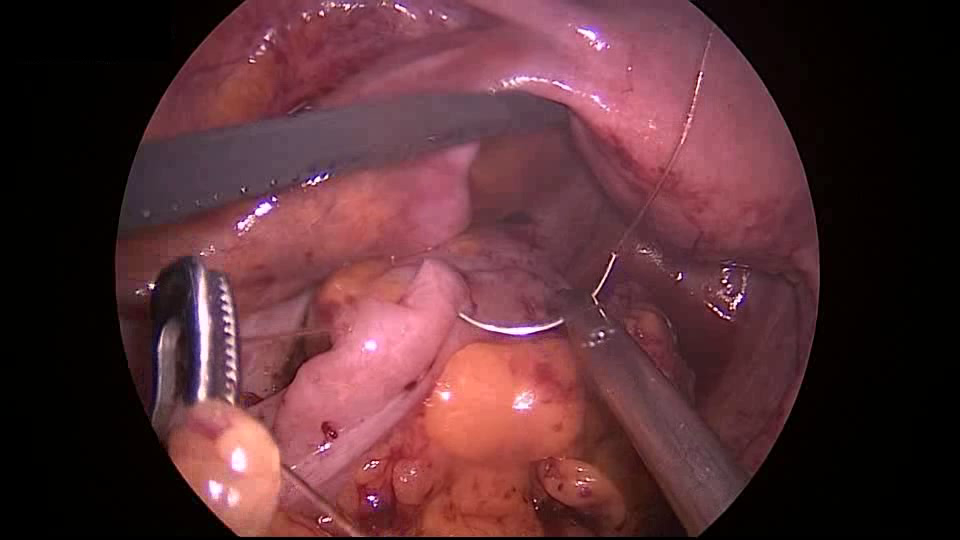}\\
\end{tabular}
\caption{Example images from minimally invasive surgeries with medical instruments. The challenges induced by motion blur~(left column), smoke~(central column), and poor illumination~(top right) are well visible. All images are taken from the ROBUST-MIS Challenge 2019 test set~\cite{ross2021comparative,maier2021heidelberg}.
}
\label{fig:exampleROBUST}
\end{figure}

The localization and pixel-precise segmentation of medical instruments in images of minimally invasive surgeries, medical instrument segmentation, is important for numerous medical applications. For instance, it allows for skill assessment~\cite{lin2019automatic}, procedure identification, or automated camera steering~\cite{zhang2020object}, among other tasks~\cite{ross2021comparative}.  However, localizing and segmenting medical instruments is limited by several application-specific characteristics~\cite{pakhomov2019deep,ross2021comparative}. Images from minimally invasive surgeries are often degraded by motion blur, blood, smoke, or poor illumination~\cite{ross2021comparative}. Some of these challenging conditions are visible in the example images in Fig.~\ref{fig:exampleROBUST}. Additionally, minimally invasive surgeries are applied in various body parts, instruments, and surrounding tissues vary. Hence, approaches for medical instrument segmentation have to be robust and generalizable to support minimally invasive surgeries in clinical practice.

Multiple formulations for medical instrument segmentation exist~\cite{allan20192017,ross2021comparative}. These include generation of a common binary mask for all instruments~\cite{laina2017concurrent,garcia2017toolnet}, segmentation of individual instances~\cite{isensee2020or,gonzalez2020isinet,ross2021comparative,ceron2021assessing} or instrument parts~\cite{shvets2018automatic,pakhomov2019deep}.
We focus on the segmentation of individual instances without classification. Various computer vision approaches based on semantic segmentation~\cite{shvets2018automatic,pakhomov2019deep,isensee2020or} and instance segmentation~\cite{gonzalez2020isinet,ross2021comparative,ceron2021assessing} were proposed for the different formulations. Besides these kinds of approaches, object proposal generation methods~\cite{Pinheiro2015-deepmask,Pinheiro2016-sharpmask,Hu2017-fastmask,WilmsFrintropACCV2018,WilmsFrintropICPR2020,WilmsFrintropIVC2021} can also be part of the armamentarium when focusing on the segmentation of individual instances. Object proposal generation methods produce a set of class-agnostic object candidates as boxes or pixel-precise masks. These candidates are typically used to limit the search space in object detection~\cite{mai2018faster} or instance segmentation~\cite{he2017mask}. Since the systems are class-agnostic, they generalize better to unseen configurations than instance segmentation methods~\cite{Pinheiro2015-deepmask,ovsep20204d,kim2021learning}.

In this paper, we adapt the object proposal generation system AttentionMask~\cite{WilmsFrintropACCV2018} to medical instrument segmentation in the context of the ROBUST-MIS Challenge 2019~\cite{ross2021comparative}. The ROBUST-MIS Challenge 2019 aims to assess the state-of-the-art in medical instrument segmentation and foster more research on this task by introducing a dataset and an evaluation framework. Utilizing AttentionMask for medical instrument segmentation is very promising since it has shown a strong performance in discovering objects of various sizes~\cite{WilmsFrintropACCV2018} and demonstrated good results in other complex real-world applications~\cite{WilmsEtAlICPR2020,wilms2022localizing}. To adapt AttentionMask, we modify the training to fit the new data and propose a dedicated post-processing to select the most relevant object proposals. The results show that our adapted AttentionMask is a promising foundation for generating high-quality medical instrument segmentations. We further investigate the influence of the instruments' size as well as the robustness and generalization ability of our system.

\section{Related Work}
\label{sec:relWork}
This section gives a brief overview of approaches to medical instrument segmentation and object proposal generation.

\subsection{Medical Instrument Segmentation}
\label{sec:relWork_mis}
As mentioned above, the different formulations of medical instrument segmentation lead to various approaches. Tackling the problem of generating one binary segmentation mask for all instruments, \cite{laina2017concurrent} and \cite{garcia2017toolnet} utilize encoder-decoder architectures. \cite{shvets2018automatic}~and \cite{pakhomov2019deep}~apply semantic segmentation networks to locate and classify different instrument parts. 

For segmenting individual instances, the task we address in this paper, several authors use instance segmentation systems. \cite{ceron2021assessing}~propose CCAM as a combination of YOLACT++~\cite{bolya2020yolact} with an attention mechanism. Most ROBUST-MIS Challenge 2019 participants that segment individual instances employ the instance segmentation system Mask R-CNN~\cite{he2017mask}. Mostly, few details like the anchor boxes or the training process are adapted to the new data. Similarly, \cite{gonzalez2020isinet}~use Mask R-CNN with temporal consistency constraints to utilize available video data. An exception to these instance segmentation-based approaches are~\cite{isensee2020or}. They first solve the binary segmentation task and extract instances using connected component analysis. 

In contrast to these approaches, we adapt an object proposal generation system showing strong generalization and robustness abilities.


\subsection{Object Proposal Generation}
\label{sec:relWork_opg}
Object proposal generation is the class-agnostic discovery of objects with boxes or masks~(proposals)~\cite{alexe2010object}. For each proposal, an objectness score is estimated to rank the proposals. Besides methods using hand-crafted features~\cite{Hosang2015PAMI}, CNN-based approaches were proposed~\cite{Pinheiro2015-deepmask,Pinheiro2016-sharpmask,Hu2017-fastmask,WilmsFrintropACCV2018,WilmsFrintropICPR2020,WilmsFrintropIVC2021}. \cite{Pinheiro2015-deepmask}~and \cite{Pinheiro2016-sharpmask} generate fore\-ground-background segmentations and objectness scores for several overlapping crops from an image pyramid. In contrast, FastMask~\cite{Hu2017-fastmask} generates a feature pyramid inside a CNN for a more efficient processing. AttentionMask~\cite{WilmsFrintropACCV2018} further improves the efficiency by utilizing attention to only extract windows in relevant parts of the feature pyramid. The improved efficiency allows a better discovery of small objects utilizing a new module. We will discuss AttentionMask in more detail in Sec.~\ref{sec:method_am} since we adapt the system to segment medical instruments.


\section{Method}
\label{sec:method}
We adapt the object proposal system AttentionMask~\cite{WilmsFrintropACCV2018} (see Fig.~\ref{fig:abstractSysFig}) to medical instrument segmentation in images of minimally invasive surgeries. As discussed above, AttentionMask showed a strong performance in localizing and segmenting objects~\cite{WilmsFrintropACCV2018} as well as a good robustness w.r.t. image degradations~\cite{WilmsEtAlICPR2020}. To adapt AttentionMask, we update the training to fit the medical instrument segmentation application and integrate a post-processing that selects the most promising object proposals in three steps. In the following, we first briefly review AttentionMask and present our updated training in Sec.~\ref{sec:method_am}. Subsequently, we introduce our dedicated post-processing in Sec.~\ref{sec:method_pp}.

\subsection{AttentionMask}
\label{sec:method_am}
\begin{figure}[t]
    \centering
    \includegraphics[width=\textwidth]{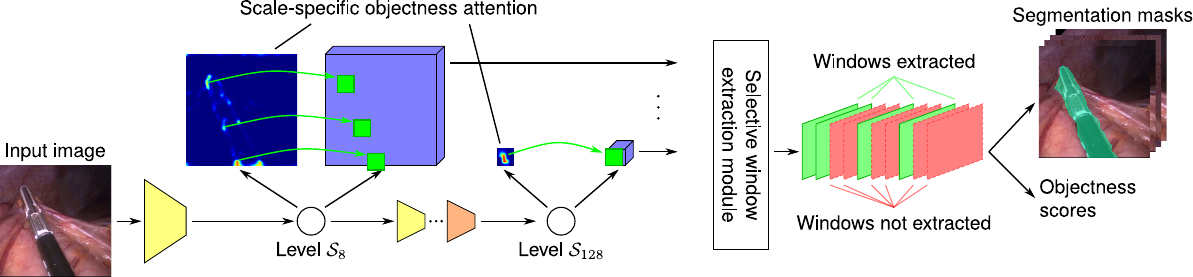}
    \caption{System figure of AttentionMask. The input image is processed by the first part of the backbone network (left yellow module), yielding the feature pyramid's base level~($\mathcal{S}_8$). Further levels are generated by downsampling up to level $\mathcal{S}_{128}$. At each pyramid level, windows are extracted in regions with high scale-specific attention (see red areas in heatmaps) to omit background areas. A pixel-precise mask and an objectness score are generated for each extracted window. Note that we cropped the input image for clarity. The figure is based on~\cite{WilmsFrintropACCV2018}.}\label{fig:abstractSysFig}
\end{figure}

AttentionMask locates and segments medical instruments by first extracting a feature map from the input image using the first part of a ResNet-based backbone~\cite{he2016deep} (left yellow module in Fig.~\ref{fig:abstractSysFig}). The result is a feature map downsampled by a factor of 8 w.r.t. the input image. This pyramid level ($\mathcal{S}_8$) serves as the base for the subsequent feature pyramid, which is generated using the remainder of the backbone and residual pooling layers (orange module in Fig.~\ref{fig:abstractSysFig})~\cite{Hu2017-fastmask}. Overall, the feature pyramid consists of 8 pyramid levels representing the image downsampled by a factor of 8 for small instruments to 128 for large instruments. 

Across the feature pyramid, several windows are extracted to locate and segment the instruments. To only extract windows in relevant areas, AttentionMask utilizes a trained attention mechanism (scale specific objectness attention in Fig.~\ref{fig:abstractSysFig}). This attention highlights areas with instruments that fit the pyramid levels in terms of size. Subsequently, $10\times 10$ windows are extracted from the feature maps in high attention areas (green windows in Fig.~\ref{fig:abstractSysFig}). Note that extracting fixed-size windows from the different pyramid levels allows AttentionMask to locate and segment instruments of various sizes. Given the extracted windows, AttentionMask generates segmentation masks for the instruments and objectness scores to rank the generated instrument proposals.

To train the system, we adapt the training regime used in~\cite{WilmsFrintropACCV2018} by reducing the learning rate to 0.00001 to fit the medical instrument segmentation data and train the system for only 12 epochs.

\subsection{Post-Processing}
\label{sec:method_pp}
We apply three post-processing steps to select the most promising 
from the initial instrument proposals. 
This step is necessary since the dense window sampling in high attention areas within AttentionMask leads to several similar proposals covering the same object. Although this is common in object proposal generation, the evaluation frameworks typically used in medical instrument segmentation penalize redundant proposals.

The first post-processing step removes instrument proposals with an objectness score below 0.8 to revoke proposals that most likely cover only object parts or background patches. Second, we combine groups of at least five partially overlapping proposals and discard all other proposals. This step counters the effects of the dense proposal generation in AttentionMask, leading to several proposals covering the same instrument with slightly different segmentations. From each group of proposals, we generate one final proposal by selecting all pixels  contained in at least 10\% of the group's proposals. These three steps remove 98.2\% of the proposals from the original AttentionMask output to better fit the evaluation frameworks commonly utilized in medical instrument segmentation. Note that we do not apply this post-processing when evaluating our approach in the object proposal generation framework.

\section{Evaluation}
\label{sec:eval}
To assess the quality of our adapted AttentionMask system, we evaluate on the dataset of the ROBUST-MIS Challenge 2019~\cite{ross2021comparative,maier2021heidelberg}. The dataset contains 10040 annotated images from 30 minimally invasive surgeries. We divide the official training set consisting of 5983 images into a training split (80\%) and a validation split (20\%). The remaining 4057 images are used as the test set similar to~\cite{ross2021comparative}. This test set was subdivided by~\cite{ross2021comparative} into three stages to test the generalization ability. While the first stage comprises images of known patients and surgeries, the second stage includes data of known surgery types but different patients. Finally, the third and most challenging stage consists of images from an unseen surgery type. Hence, the second and third stages of the test data are independent of the training data. We follow~\cite{ross2021comparative} and report the stage-specific results.

For measuring the quality of the results, we use two strategies. First, we follow the ROBUST-MIS Challenge 2019~\cite{ross2021comparative} and calculate the Multi-instance Dice Similarity Coefficient (MI\_DSC) and the Multi-instance Normalized Surface Dice (MI\_NSD). While the MI\_DSC assesses the segmentations' quality by comparing entire regions with the annotation, MI\_NSD only uses the areas around the contours. To better assess the consistency of the performance and the applicability in clinical practice, we focus on the worst-case results and report the 5\% quantile for MI\_DSC and MI\_NSD as in~\cite{ross2021comparative}. 

As a second evaluation strategy, we use a standard object proposal evaluation protocol~\cite{Pinheiro2015-deepmask,Pinheiro2016-sharpmask,Hu2017-fastmask,WilmsFrintropACCV2018,WilmsFrintropICPR2020,WilmsFrintropIVC2021}. Hence, we evaluate the results using Average Recall (AR)~\cite{Hosang2015PAMI} for different numbers of proposals. AR assesses how many annotated objects are discovered and how well they are segmented. We also calculate the recall for annotated objects of different size categories, similar to~\cite{Pinheiro2015-deepmask,Pinheiro2016-sharpmask,Hu2017-fastmask,WilmsFrintropACCV2018,WilmsFrintropICPR2020,WilmsFrintropIVC2021}. The recall defines how many annotated objects are located given an Intersection over Union (IoU) value for assessing the overlap between the proposal and the annotated object. We define our size categories in terms of relative size (object area divided by image area). The categories are XS ($<1\%$), S($1\%-2\%$), M ($2\%-5\%$), L ($5\%-10\%$), and XL ($>10\%$).

We compare our results to the participants of the ROBUST-MIS Challenge 2019 \textit{caresyntax}, \textit{CASIA\_SRL}, \textit{fisensee}~\cite{isensee2020or}, \textit{SQUASH}, \textit{Uniandes}, \textit{VIE}, and \textit{www} based on the results presented in~\cite{ross2021comparative}. Additionally, we compare to CCAM~\cite{ceron2021assessing} in the evaluation framework of~\cite{ross2021comparative}. Note that we evaluate different versions of our proposed approach. In the ROBUST-MIS Challenge 2019 evaluation framework~(Sec.~\ref{sec_eval_quan_mis}), we evaluate our adapted AttentionMask with and without the dedicated post-processing as well as the adapted AttentionMask with an optimal ranking. The latter version gives an upper limit for our approach and selects the best fitting proposal per ground truth object. Hence, it assumes an optimal ranking. In the object proposal generation framework~(Sec.~\ref{sec_eval_quan_opg} and Sec.~\ref{sec_eval_qual}), we only evaluate our adapted AttentionMask without the dedicated post-processing to match the evaluation framework.


\subsection{Quantitative Results in the ROBUST-MIS Framework}
\label{sec_eval_quan_mis}
First, we present the results using MI\_DSC and MI\_NSD on the ROBUST-MIS Challenge 2019 test set stage 3 in Tab.~\ref{tab:dsc_nsd}. This test set only features a surgery type not seen in training. The results indicate that our dedicated post-processing improves the results of our adapted AttentionMask system, but only outperforms two and three other systems, respectively. The best systems in terms of MI\_DSC outperform our system with post-processing by 0.17 due to the impaired initial ranking. If we assume an optimal ranking for our system, the results substantially improve, outperforming all systems by at least 67.8\% and 74.3\% in terms of MI\_DSC and MI\_NSD. These results show the potential of our adapted AttentionMask system for medical instrument segmentations, but also clearly indicate the low quality of the initial ranking. 

Note that a result of 0.00 does not imply that a method misses all instruments, since we only focus on the worst-case scenarios and report 5\% quantiles~\cite{ross2021comparative}.

\begin{table}[t]
\centering
\caption{Quantitative results of the three versions of our adapted AttentionMask, CCAM~\cite{ceron2021assessing}, and the participants of the ROBUST-MIS Challenge 2019~\cite{ross2021comparative} in terms of Multi-instance Dice Similarity Coefficient (MI\_DSC) and Multi-instance Normalized Surface Dice (MI\_NSD) on stage 3 of the ROBUST-MIS Challenge 2019 test set~\cite{ross2021comparative,maier2021heidelberg}. All results are 5\% quantiles. \textbf{Bold font} highlights the best current results, while \textit{italic font} denotes the best results given an optimal ranking for our system to showcase its potential.}
\label{tab:dsc_nsd}
\begin{tabular}{lcc}
\toprule
Participant/System & MI\_DSC$\uparrow$ & MI\_NSD$\uparrow$ \\ \midrule
VIE &  0.00 & 0.00 \\
caresyntax &  0.00 & 0.00 \\
fisensee~\cite{isensee2020or} &  0.17 & 0.16 \\
CASIA\_SRL &  0.19 & 0.27 \\
SQUASH &  0.22 & 0.26 \\
Uniandes &  0.26 & 0.29 \\
www &  \textbf{0.31} & \textbf{0.35} \\ \midrule
CCAM \cite{ceron2021assessing} &  \textbf{0.31} & 0.34 \\ \midrule
Ours~(without post-processing) & 0.00 & 0.00 \\
Ours~(with post-processing) & 0.14 & 0.19 \\ 
Ours~(with optimal ranking) & \textit{0.52} & \textit{0.61} \\ \bottomrule
\end{tabular}
\end{table}

\subsection{Quantitative Object Proposal Generation Results}
\label{sec_eval_quan_opg}
In the second part of our evaluation, we analyze our adapted AttentionMask without the dedicated post-processing in more detail using Average Recall (AR) measures. Removing the post-processing matches the evaluation framework of object proposal generation. Table~\ref{tab:robustmis_ar} presents the overall results in terms of AR for the three stages of the ROBUST-MIS Challenge 2019 test set. The results indicate that the first 10 proposals locate most instruments that our adapted AttentionMask discovers. Between AR@1 and AR@10, the results improve by 0.280 across all stages, while the improvement from AR@10 to AR@100 is only 0.063 on average. Hence, the rough ranking of proposals in AttentionMask has a high quality, while the precise ranking of the most relevant proposals is suboptimal. Comparing the results across the different stages, no major difference is visible. For instance, in terms of AR@10 the difference between the best result (stage 2) and the worst result (stage 3) is only 0.082. Therefore, AttentionMask shows strong generalization abilities across different patients (stage 2) and surgery types (stage 3). 

\begin{table}[t]
\centering
\caption{Quantitative results of our adapted AttentionMask without post-processing in terms of Average Recall (AR) using different numbers of proposals on the three stages of the ROBUST-MIS Challenge 2019 test set~\cite{ross2021comparative,maier2021heidelberg}.}
\label{tab:robustmis_ar}
\begin{tabular}{lccc}
\toprule
Test split stage & AR@1$\uparrow$ & AR@10$\uparrow$ & AR@100$\uparrow$ \\ \midrule
Stage 1 &  0.182 & 0.471 & 0.533 \\
Stage 2 &  0.214 & 0.502 & 0.554 \\
Stage 3 &  0.156 & 0.420 & 0.497 \\ \bottomrule
\end{tabular}
\end{table}

\begin{figure}
\centering
\subcaptionbox{Stage 1\label{fig:recSize_stage1}}{\includegraphics[width =0.32\linewidth]{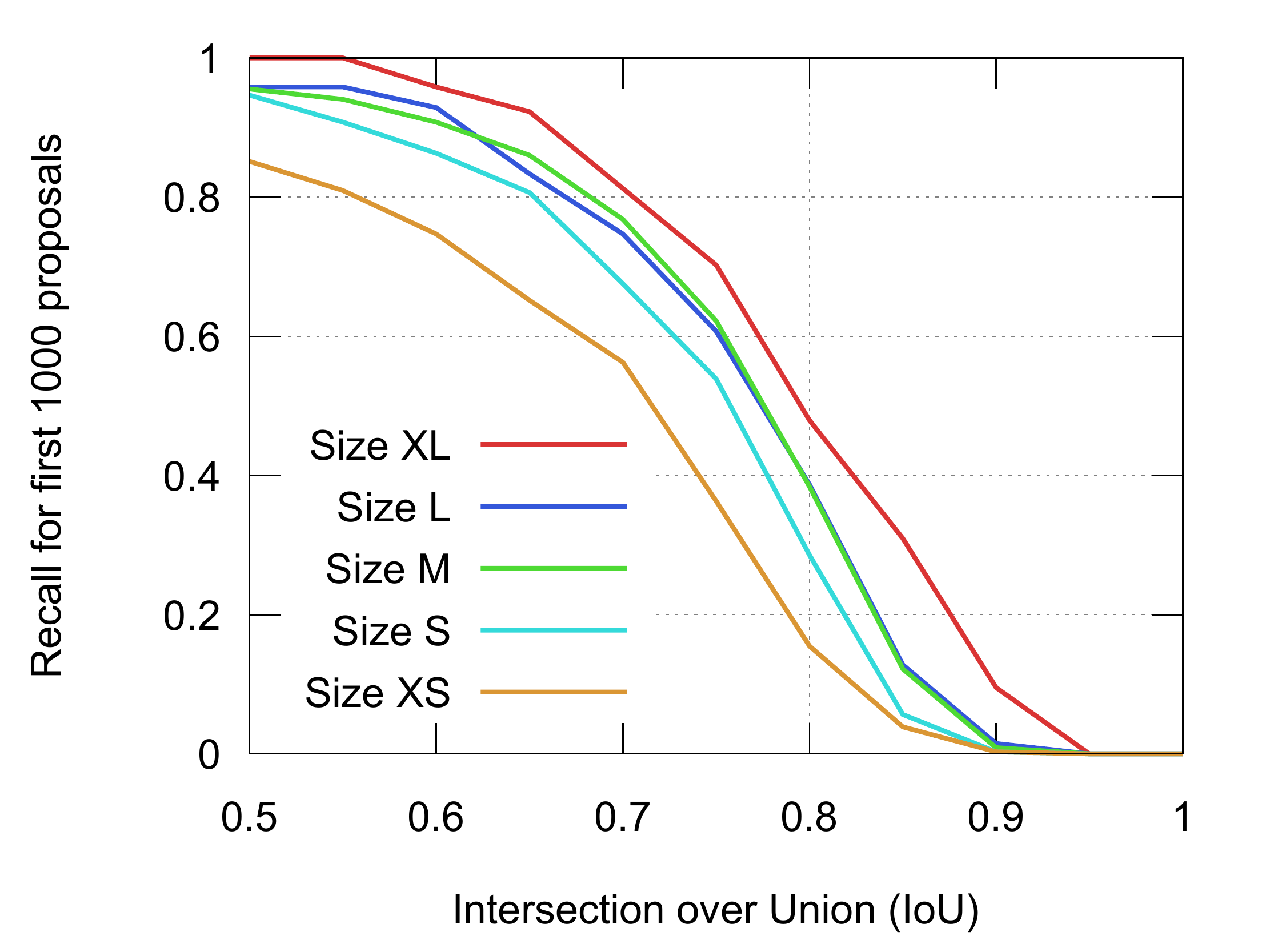}}%
\hfill
\subcaptionbox{Stage 2\label{fig:recSize_stage2}}{\includegraphics[width =0.32\linewidth]{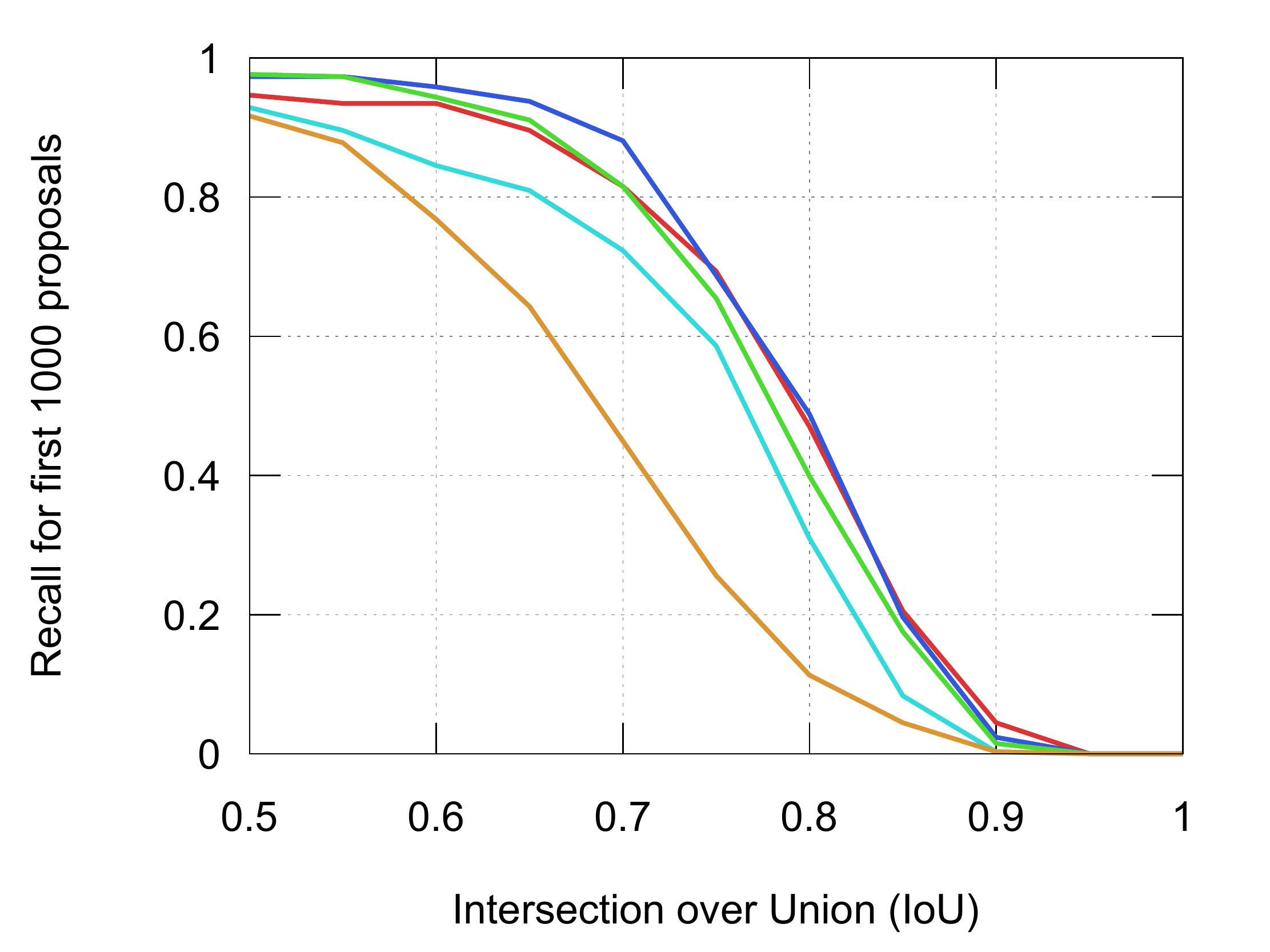}}%
\hfill
\subcaptionbox{Stage 3\label{fig:recSize_stage3}}{\includegraphics[width =0.32\linewidth]{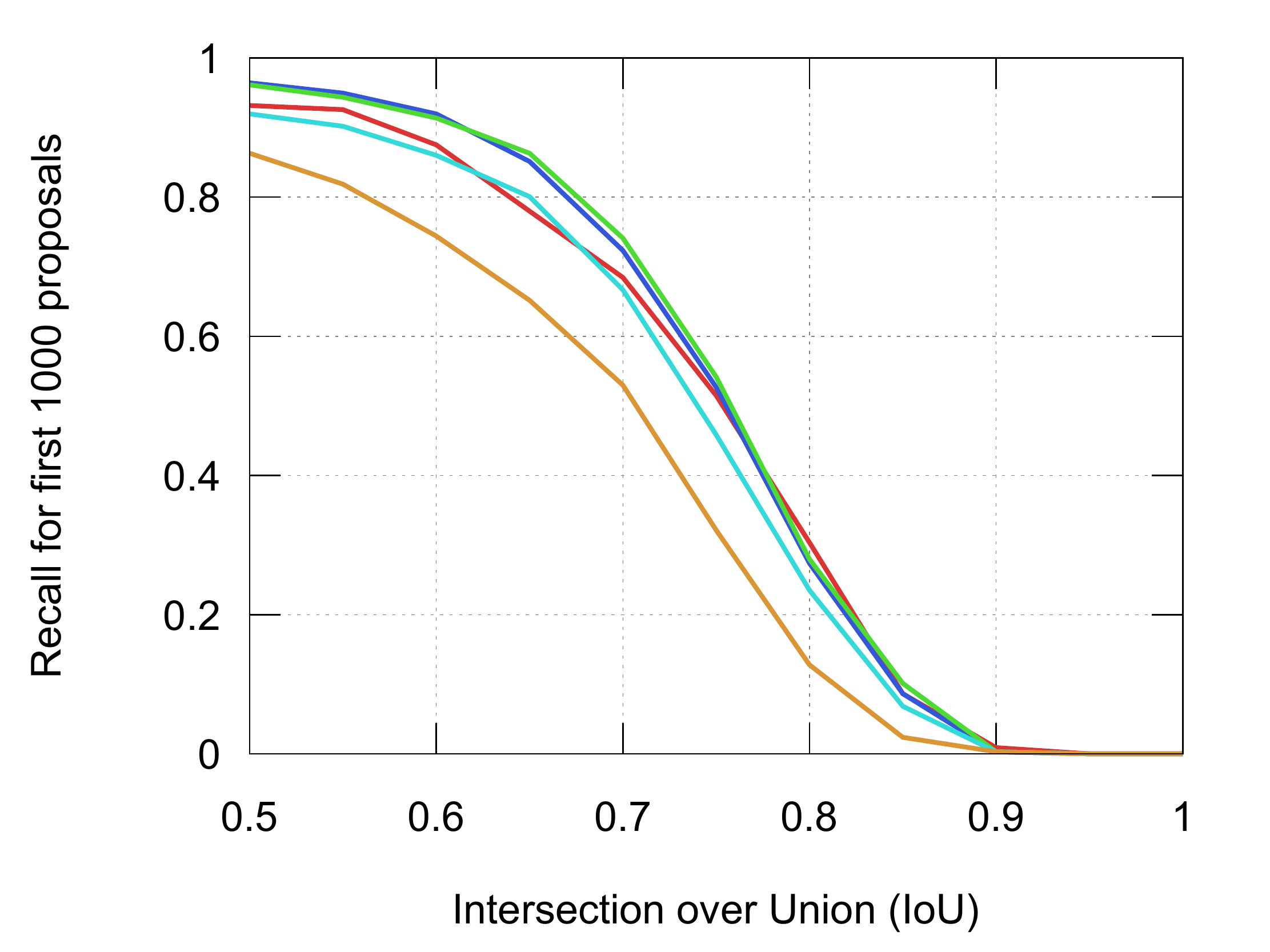}}%
\caption{Results of our adapted AttentionMask without post-processing in terms of recall on the three stages of the ROBUST-MIS Challenge 2019 test set~\cite{ross2021comparative,maier2021heidelberg}. The results per stage are divided by the relative size of the instruments. The categories are XS ($<1\%$), S($1\%-2\%$), M ($2\%-5\%$), L ($5\%-10\%$), and XL ($>10\%$).
}
\label{fig:robust_misc_recSize}
\end{figure}

We also evaluate the quality of our instrument segmentations w.r.t. the instruments' size. Figure~\ref{fig:robust_misc_recSize} shows the results per stage and size category in terms of the located instruments~(recall) for different Intersection over Union (IoU) levels. The results indicate that except for very small instruments (XS), the results for the different size categories (S-XL) are very similar. However, even for very small instruments, more than 80\% of the instruments are located and segmented properly using the common IoU threshold of 0.5~\cite{Everingham10}. These results are in line with the findings in~\cite{WilmsFrintropACCV2018}, showing a strong performance of AttentionMask on small objects. Only for IoU levels above 0.7 the results across all sizes start to drop substantially. Overall, the results of AttentionMask are only weakly related to the instruments' size, which is different from the participants of the ROBUST-MIS Challenge 2019 according to~\cite{ross2021comparative}.

\subsection{Qualitiative Results}
\label{sec_eval_qual}
\begin{figure}[t]
\centering
\begin{tabular}{ccc}
\includegraphics[width = .31\linewidth]{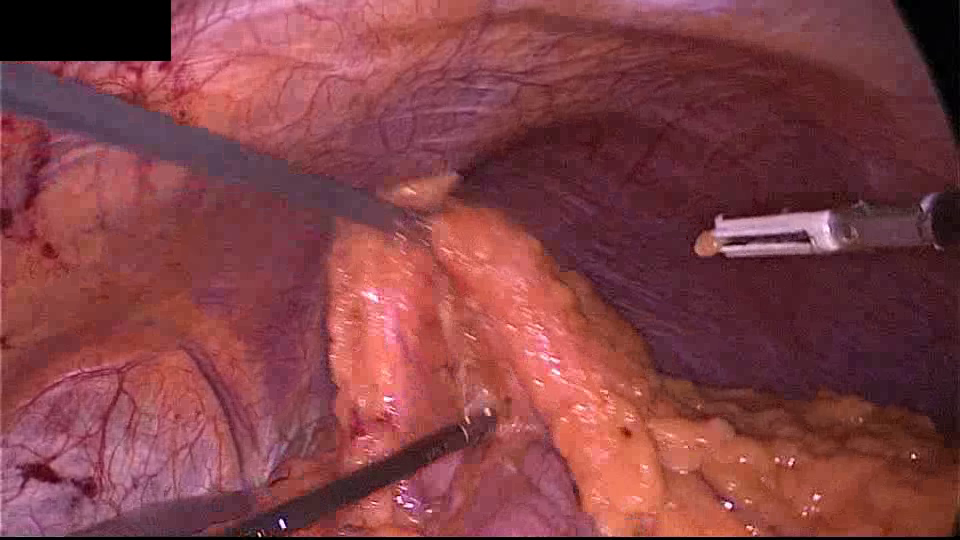} &
\includegraphics[width = .31\linewidth]{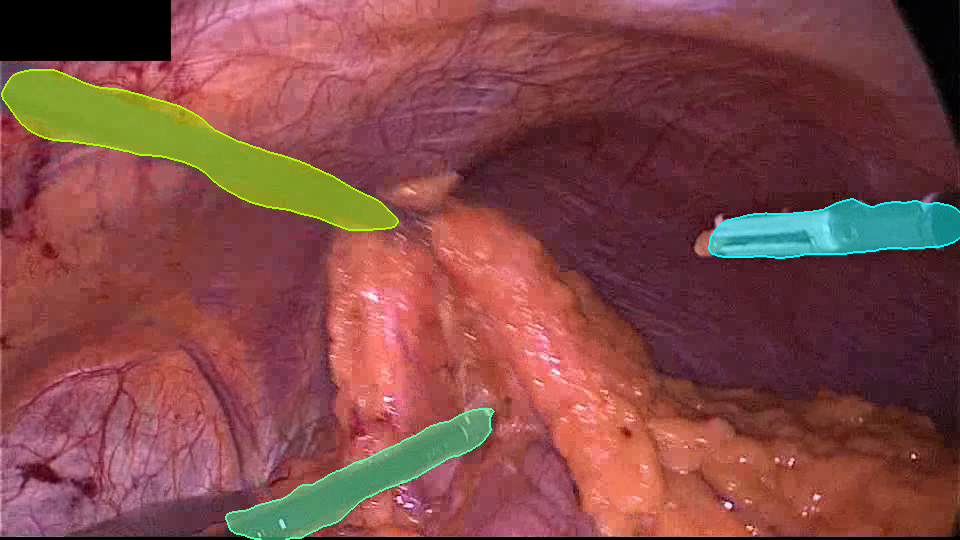} &
\includegraphics[width = .31\linewidth]{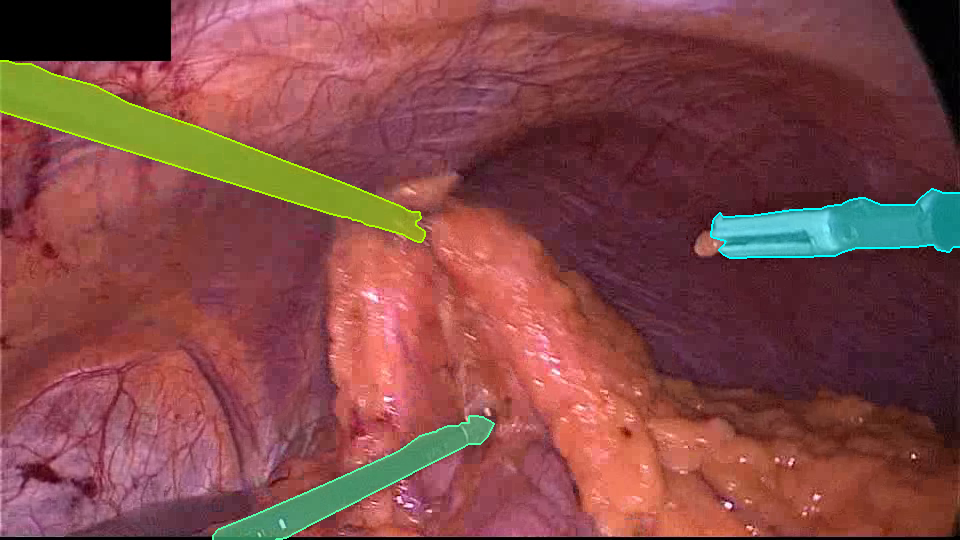}\\
\includegraphics[width = .31\linewidth]{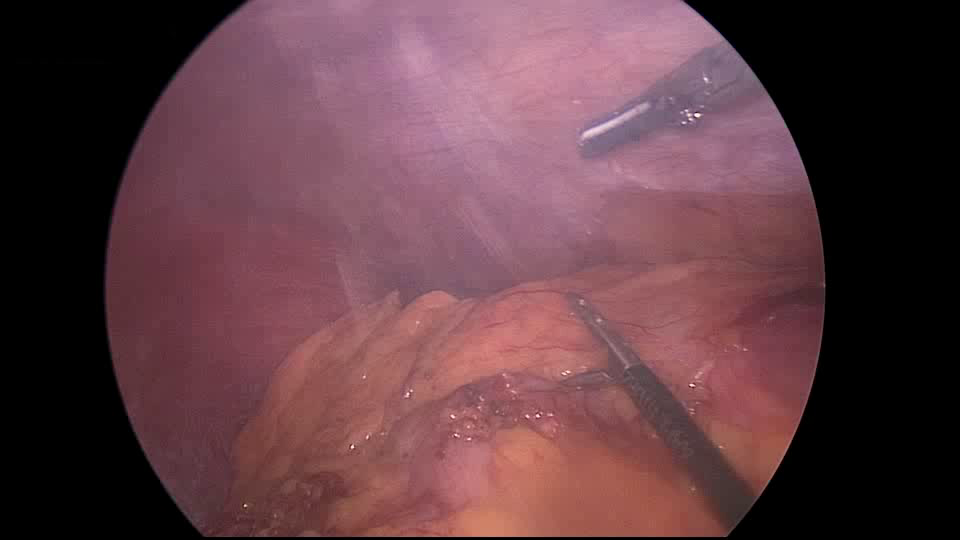} &
\includegraphics[width = .31\linewidth]{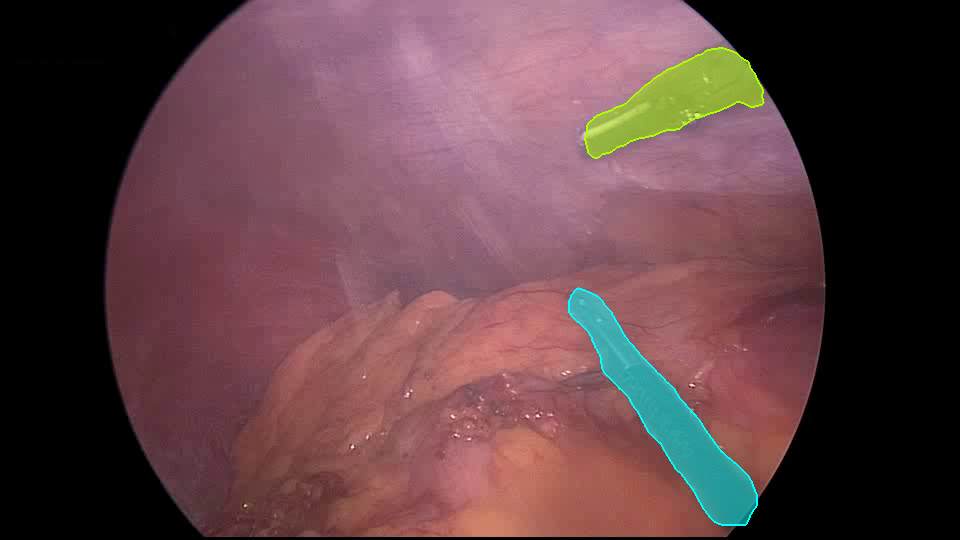} &
\includegraphics[width = .31\linewidth]{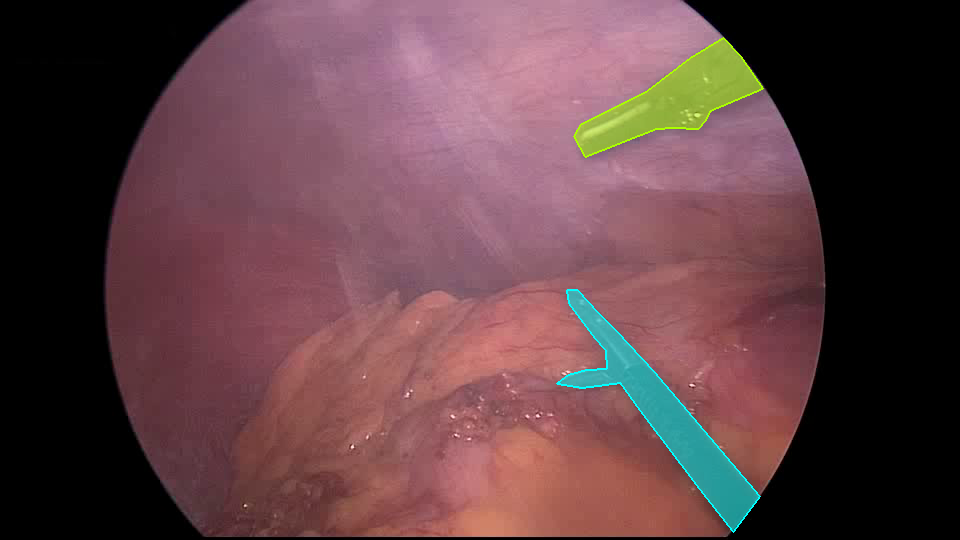}\\
\includegraphics[width = .31\linewidth]{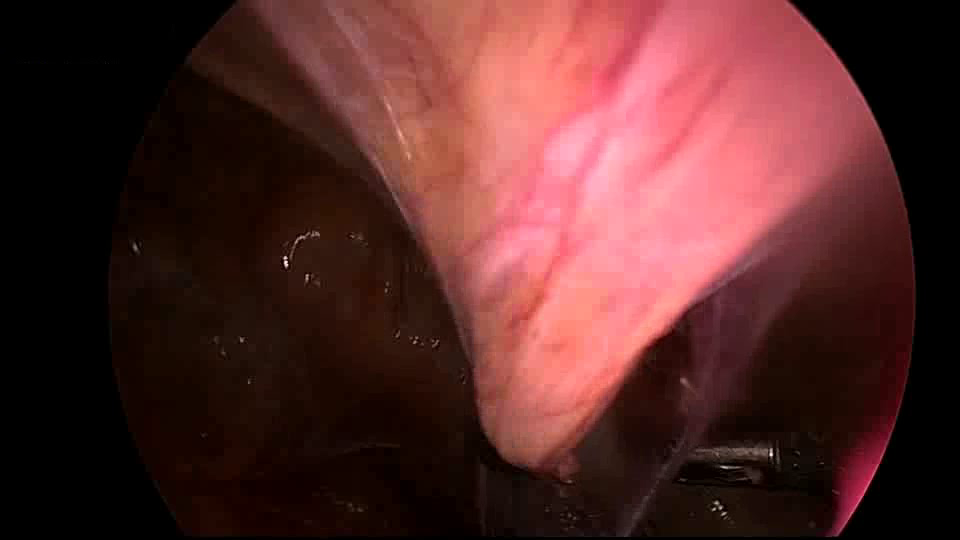} &
\includegraphics[width = .31\linewidth]{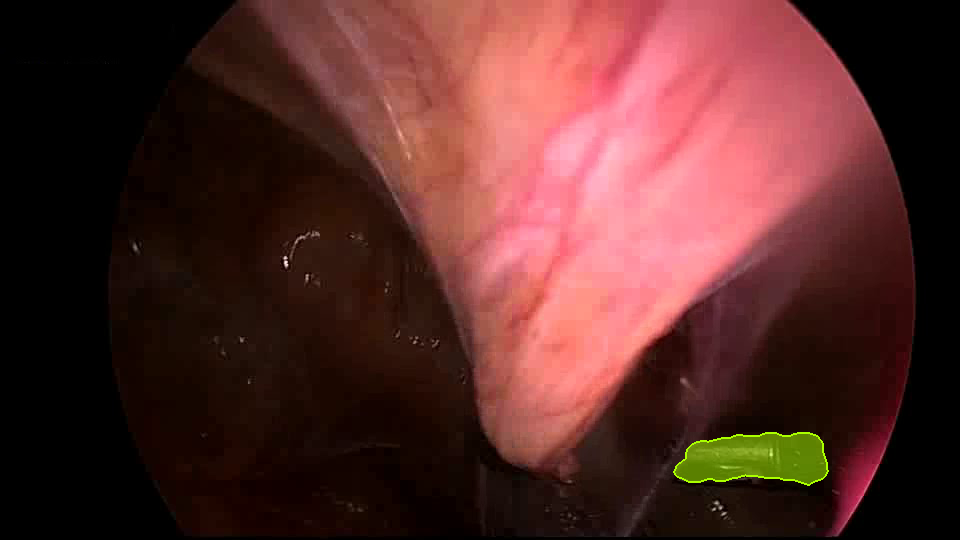} &
\includegraphics[width = .31\linewidth]{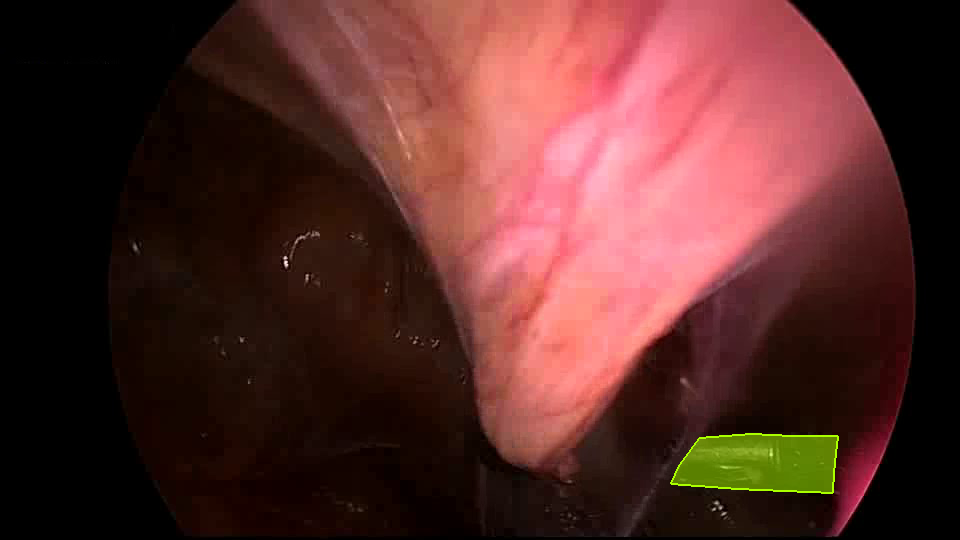}\\
\includegraphics[width = .31\linewidth]{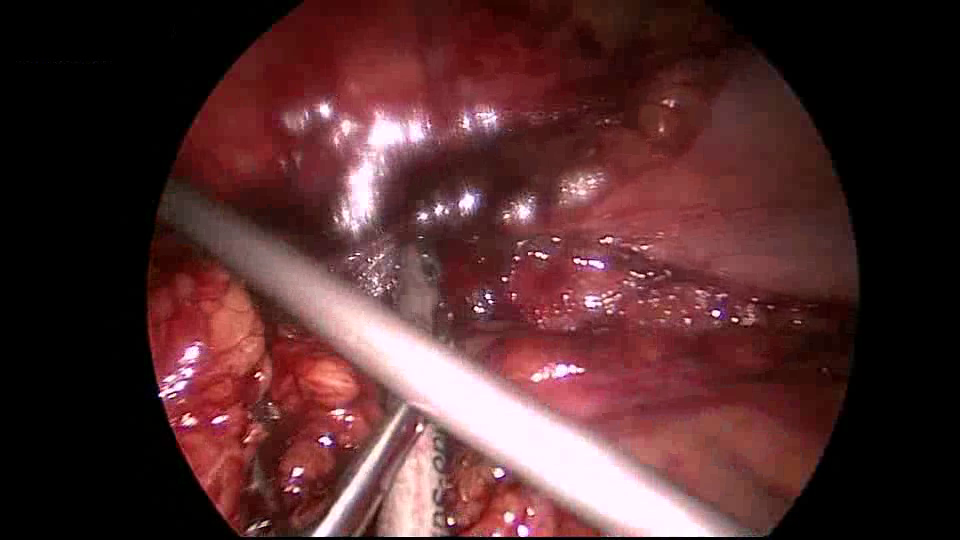} &
\includegraphics[width = .31\linewidth]{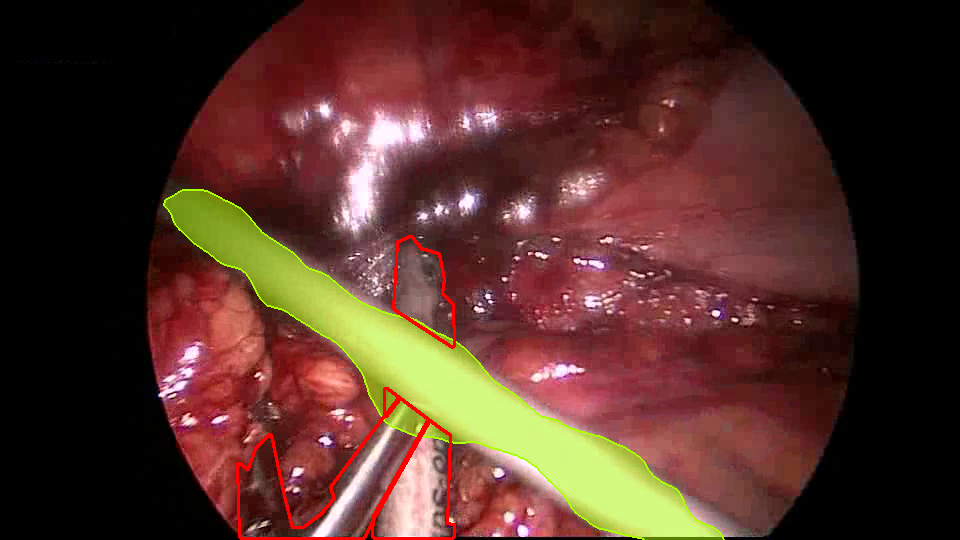} &
\includegraphics[width = .31\linewidth]{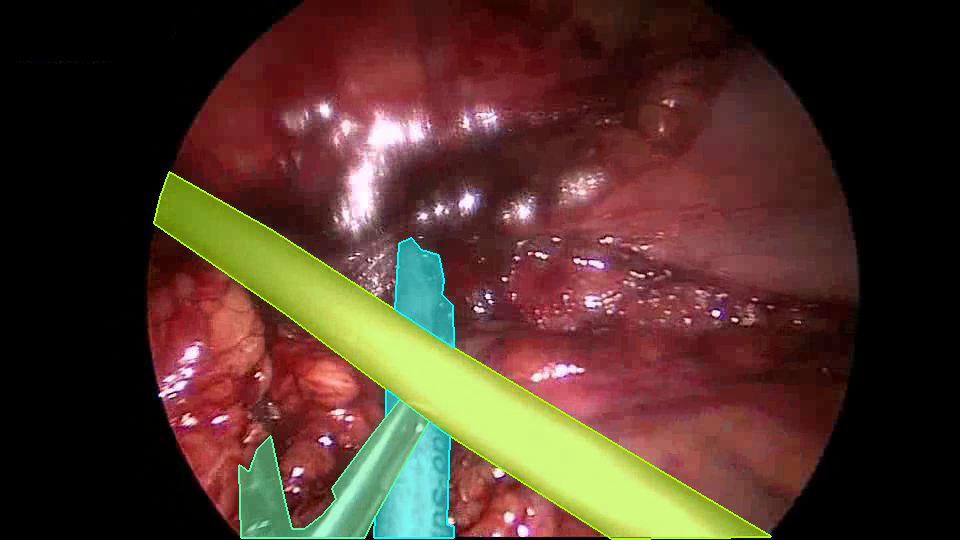}\\
{\footnotesize Input image} &  {\footnotesize Ours} & {\footnotesize Ground truth}
\end{tabular}
\caption{Qualitative results of our adapted AttentionMask without post-processing on four images from the ROBUST-MIS Challenge 2019 test set~\cite{ross2021comparative,maier2021heidelberg} covering different challenging conditions. Filled colored contours denote found objects, while not filled red contours denote missed objects. Note that only the best fitting proposal (highest IoU) is visualized per annotated object.
}
\label{fig:qualResROBUST}
\end{figure}

Figure~\ref{fig:qualResROBUST} presents qualitative results of our adapted AttentionMask on four images from the challenging stage 3 of the ROBUST-MIS Challenge 2019 test set. The first three examples show high-quality instrument segmentations, despite different challenges like multiple objects, smoke, or suboptimal illumination. These results indicate a strong robustness of AttentionMask and are in line with the findings in~\cite{ross2021comparative} and~\cite{WilmsEtAlICPR2020}. However, the final example shows a challenging scenario where AttentionMask fails to discover two of three instruments. We attribute this error to the strong occlusion of the green and blue instruments (mask color in ground truth). Another typical limitation is visible in the second example. Due to the inherited downsampling process in CNNs, AttentionMask misses small details of the instruments, like parts of the tip in this example. This behavior is similar to the findings of~\cite{WilmsFrintropICPR2020,WilmsFrintropIVC2021} applying AttentionMask on typical computer vision datasets.

\section{Conclusion}
\label{sec:conclusion}
This paper addressed medical instrument segmentation in minimally invasive surgery images. We adapted the object proposal generation system AttentionMask and proposed a dedicated post-processing pipeline to utilize AttentionMask for medical instrument segmentation. Results on the ROBUST-MIS Challenge 2019 dataset showed that our adapted AttentionMask system is a promising foundation for precise and robust medical instrument segmentation. Specifically, we showed robustness of our system w.r.t. image degradations like smoke or suboptimal illumination. Importantly, AttentionMask-based medical instrument segmentation generalizes well to unseen scenarios and localizes instruments of almost all sizes with high quality. The latter behavior is especially intriguing since it differs from previous systems~\cite{ross2021comparative}.

A main limitation of our system lies in the non-optimal ranking of the segmentation mask proposals. As some of these proposals do not accurately reflect the detailed shape of the corresponding object, non-optimal ranking results in a considerable performance degradation in terms of segmentation metrics that focus exactly on the quality of the segmented object outline. We have shown that falsely high ranking of inaccurate proposals is the central hurdle for our system to outperform state-of-the-art systems. Superpixel-based refinement~\cite{WilmsFrintropICPR2020,WilmsFrintropIVC2021} may alleviate this problem through reduction of ill-shaped object proposals.


\bibliographystyle{splncs04}
\bibliography{lit}
\end{document}